\pdfoutput=1

\documentclass[11pt]{article}

\usepackage[final]{acl}

\usepackage{times}
\usepackage{latexsym}

\usepackage[T1]{fontenc}

\usepackage[utf8]{inputenc}

\usepackage{microtype}

\usepackage{inconsolata}
\usepackage[textsize=tiny]{todonotes}

\usepackage{graphicx}
\usepackage{multirow}
\usepackage{multicol}
\usepackage{tabularx}
\usepackage{amssymb}
\usepackage{mathtools}
\usepackage{enumitem}
\usepackage{amsthm}
\usepackage{bbm}
\usepackage{comment}
\usepackage{caption}
\usepackage{subcaption}
\usepackage{wrapfig}
\usepackage{pifont}
\usepackage{tabularx}
\usepackage{booktabs}
\usepackage{colortbl}
\usepackage{url}
\usepackage{algorithm}
\usepackage{algpseudocode}

\usepackage[skins,breakable]{tcolorbox}

\newcommand{\headname}{\textsc{QRHead}}
\newcommand{\headsname}{\textsc{QRHead}}
\newcommand{\methodname}{\textsc{QRRetriever}}
\newcommand{\worsehead}{\textsc{RetHead}}
\newcommand{\retscore}{\mathcal{R}}
\newcommand{\clipper}{\textsc{Clipper}}

\newcommand{\STAB}[1]{\begin{tabular}{@{}c@{}}#1\end{tabular}}

\definecolor{light-purple}{RGB}{151,156,171}
\definecolor{blue-color}{RGB}{40,166,189}
\definecolor{pink-color}{RGB}{237,46,104} 
\definecolor{dark-grey-color}{RGB}{79,91,102}

\newtcolorbox[list inside=prompt,auto counter,number within=section]{prompt}[1][]{
    colbacktitle=black!80,
    colframe=black!80,
    coltitle=white,
    fontupper=\footnotesize,
    boxsep=5pt,
    left=0pt,
    right=0pt,
    top=0pt,
    bottom=0pt,
    boxrule=1pt,
    enhanced, 
    breakable,
    skin first=enhanced,
    skin middle=enhanced,
    skin last=enhanced,
    #1,
}

\definecolor{exsinputcolor}{HTML}{E4F2DA}
\definecolor{exsoutputcolor}{HTML}{EEE2FB}
\newtcolorbox[list inside=prompt,auto counter,number within=section]{exsinput}[1][]{
    colback=exsinputcolor,
    colbacktitle=black!80,
    colframe=black!80,
    coltitle=white,
    fontupper=\footnotesize,
    boxsep=5pt,
    left=0pt,
    right=0pt,
    top=0pt,
    bottom=0pt,
    boxrule=1pt,
    enhanced, 
    breakable,
    skin first=enhanced,
    skin middle=enhanced,
    skin last=enhanced,
    #1,
}

\newtcolorbox{exsoutput}[1][]{
    colback=exsoutputcolor,
    colbacktitle=black!80,
    colframe=black!80,
    coltitle=white,
    fontupper=\footnotesize,
    boxsep=5pt,
    left=0pt,
    right=0pt,
    top=0pt,
    bottom=0pt,
    boxrule=1pt,
    enhanced, 
    breakable,
    skin first=enhanced,
    skin middle=enhanced,
    skin last=enhanced,
    #1,
}

\title{Query-Focused Retrieval Heads Improve \\ Long-Context Reasoning and Re-ranking}

\author{
 Wuwei Zhang$^\spadesuit$ \quad  Fangcong Yin$^\diamondsuit$ \quad  Howard Yen$^\spadesuit$ \quad Danqi Chen$^\spadesuit$ \quad Xi Ye$^\spadesuit$ \\
 $^{\spadesuit}$ Princeton Language and Intelligence, Princeton University \\
 $^{\diamondsuit}$ The University of Texas at Austin \\
   $^{\spadesuit}$ {\texttt{\{wuwei.zhang,hyen,danqic\}@cs.princeton.edu} \quad xi.ye@princeton.edu } \\
   $^{\diamondsuit}$ {\texttt{fangcongyin@utexas.edu} } \\
}

\begin{document}

\maketitle

\begin{abstract}

Recent work has identified retrieval heads~\cite{wu2025retrieval}, a subset of attention heads responsible for retrieving salient information in long-context language models (LMs), as measured by their copy-paste behavior in Needle-in-a-Haystack tasks.
In this paper, we introduce \headname{} (\underline{Q}uery-Focused \underline{R}etrieval Head), an improved set of attention heads that enhance retrieval from long context. We identify \headname{} by aggregating attention scores  \emph{with respect to the input query}, using a handful of examples from real-world tasks (e.g., long-context QA).
We further introduce \methodname{}, an efficient and effective retriever that uses the accumulated attention mass of \headname{} as retrieval scores.
We use \methodname{} for long-context reasoning by selecting the most relevant parts with the highest retrieval scores. On multi-hop reasoning tasks LongMemEval and \clipper{}, this yields over 10\% performance gains over full context and outperforms strong dense retrievers.
We also evaluate \methodname{} as a re-ranker on the BEIR benchmark and find that it achieves strong zero-shot performance, outperforming other LLM-based re-rankers such as RankGPT.
Further analysis shows that both the query-context attention scoring and task selection are crucial for identifying \headname{} with strong downstream utility.
Overall, our work contributes a general-purpose retriever and offers interpretability insights into the long-context capabilities of LMs.%
\footnote{Code: \url{https://github.com/princeton-pli/QRHead}.}

\end{abstract}

\section{Introduction}

Retrieving salient information from long context serves as a foundation for language models (LMs), enabling a wide range of downstream applications, such as long document understanding and passage re-ranking.
Prior work has identified a subset of attention heads in transformers~\cite{Vaswani+2017} that are responsible for retrieving relevant information, known as \emph{retrieval heads} \citep{wu2025retrieval}.

\begin{figure}[t]
    \centering
    \includegraphics[width=0.85\linewidth,trim=0 10 0 0,clip]{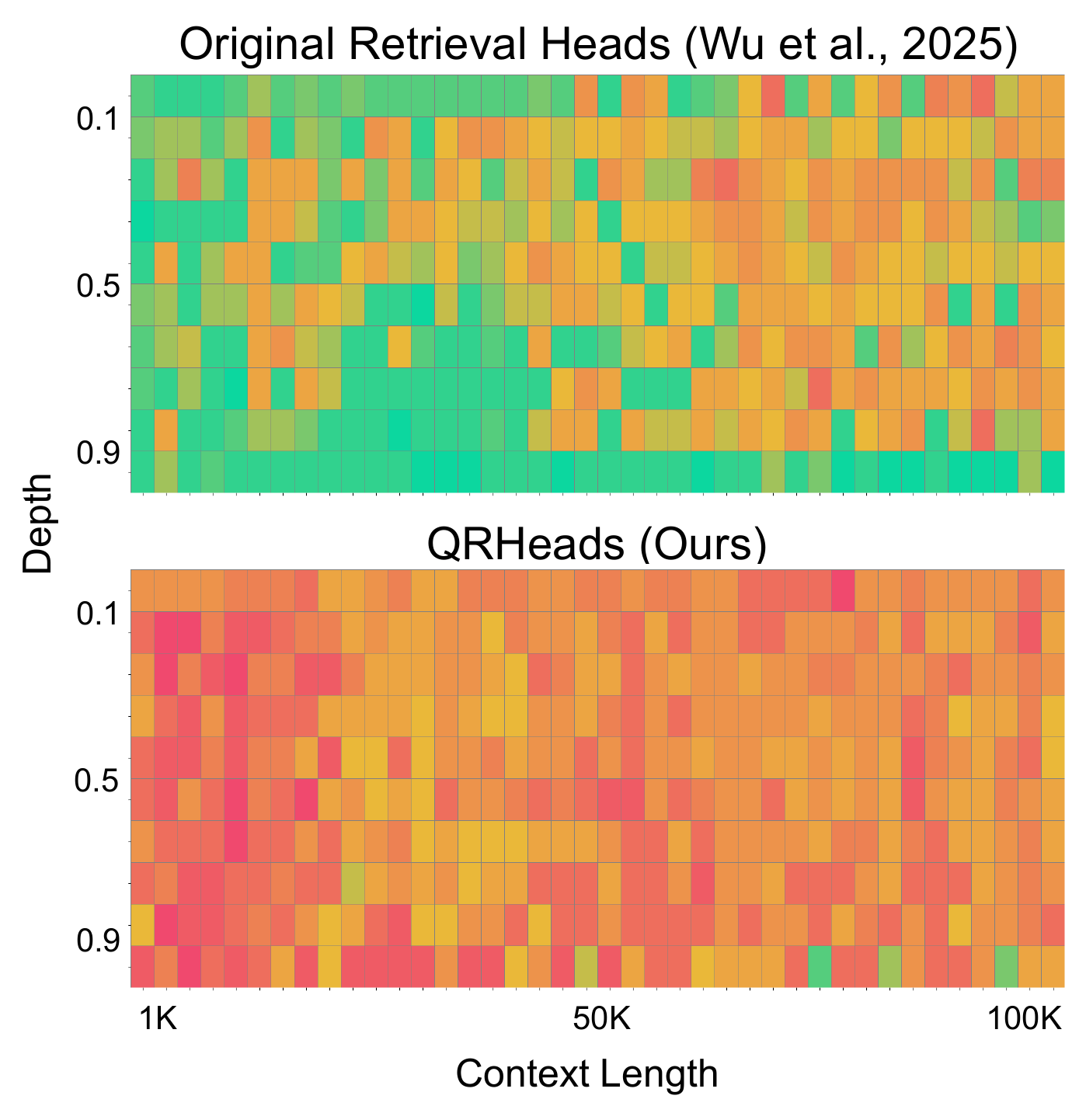}
    \vspace{-0.5em}
    \caption{
    \textbf{Top}: Masking the top 32 original retrieval heads~\cite{wu2025retrieval} of Llama-3.1-8B. \textbf{Bottom}: Masking the top 32 QRHeads of the same model, which has a more pronounced impact on Needle-in-a-Haystack.}
    \vspace{-1em}
    \label{fig:niah_teaser}
\end{figure}

However, these retrieval heads are identified based on the frequency of their copy-paste operations in a simple synthetic task—Needle-in-a-Haystack ~\citep[NIAH;][]{gkamradt_llmtest_needleinahaystack_2024}.
Although they exhibit significance on certain downstream tasks, such as extractive question answering, we argue that the copy-paste objective and synthetic data used to identify them are misaligned with how language models retrieve pertinent information in real-world settings.

To this end, we propose a more effective approach for identifying retrieval heads and introduce \headname{}, a distinct subset of attention heads whose attention mass plays a more critical role in retrieving relevant information from long context.
Compared to original retrieval heads, our method incorporates two key changes:
(1) a query-context scoring function that measures attention mass allocated to pertinent context spans with respect to an input query, and
(2) the use of more natural data from real-world tasks, such as question answering over long texts. Our method only requires a small amount of data to be effective. As shown in Figure~\ref{fig:niah_teaser}, we detect \headname{} using 70 examples from a natural long-context QA task, LongMemEval, and find masking them out results in more severe degradation in NIAH compared to original retrieval heads detected from in-domain data.

Furthermore, we build \methodname{} on top of \headname{} as a general-purpose retriever for improving LMs on diverse long-context downstream applications.
Given a query and a set of passages (e.g., a claim and a book consisting of multiple chapters), \methodname{} scores each passage using the aggregated attention mass from the \headname{} of a language model, and returns the top-ranked passages. We detect \headsname{} for multiple LMs of different scales (3B–70B) and families (Llama-3.1, Llama-3.2, and Qwen), and build \methodname{} with these LLMs.

We evaluate \methodname{} on two long-context, multi-hop reasoning tasks: LongMemEval~\cite{wu2025longmemeval} and \clipper{}~\cite{pham2025clipper}. Using \methodname{} to select top-ranked documents yields substantial improvements in retrieval recall and downstream task performance. For example, with Llama-3.1-8B-Instruct, \methodname{} outperforms dense retrievers and improves performance by over 10\% on both datasets, compared to full-context generation.
We further evaluate \methodname{} as a re-ranker on the standard BEIR benchmark~\cite{thakur2021beir}. It exhibits strong zero-shot performance across diverse domains and outperforms other LLM-based re-rankers, such as RankGPT~\cite{sun2024rankgpt}.

Finally, we provide extensive analyses of the effectiveness of \headname{}. First, using \headname{} outperforms both full attention heads and original retrieval heads. Second, \headname{} \textbf{generalizes across different tasks and input lengths}—the heads identified at 32K tokens transfer well to tasks with 128K context lengths. Lastly, we show that both key modifications—our query-focused scoring objective and the use of natural data—contribute to the improved downstream performance of \headname{} over original retrieval heads. Together, these findings highlight the practicality and robustness of \headname{} as a foundation for long-context retrieval and suggest opportunities for further exploration of retrieval mechanisms in language models.

\begin{figure*}
\centering
\begin{subfigure}{.85\textwidth}
  \centering
  \includegraphics[width=1\linewidth]{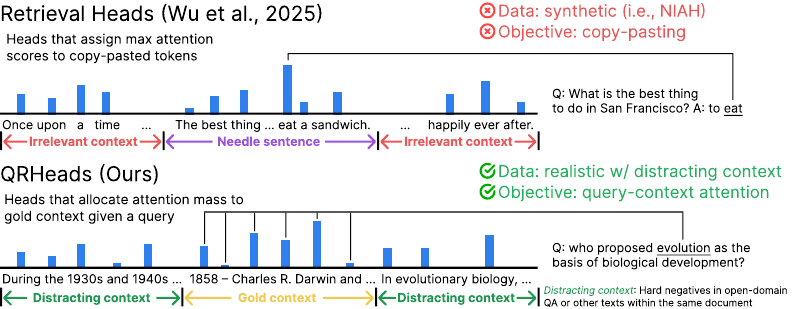}
\end{subfigure}%
\vspace{-0.2em}
\caption{
Comparison between Retrieval Heads \citep{wu2025retrieval} and \headsname{} (Ours).
}  \label{fig:methods}
\end{figure*}

\section{Background: Retrieval Heads}
\label{sec:background}

Retrieval heads are a specialized subset of attention heads that are pivotal for extracting relevant information from long input context. 

\paragraph{Original retrieval heads.}
\citet{wu2025retrieval} first discovered a set of retrieval heads that exhibit copy-paste behavior during decoding---effectively copying tokens from the long context input context into the generated output. As shown in Figure~\ref{fig:methods} (top), the retrieval head detection method roots from the Needle-in-a-Haystack test (NIAH) with a triple $(C,q,a)$ of context, question, and answer: the answer span $a$ (the ``needle'') is embedded within a long context sequence $C=d_{1}...a...d_{N}$ where $d_{1},...,d_{N}$ are $N$ irrelevant passages (the ``haystack''). 
The LM is tasked to generate an answer to $q$ based on the provided context. Successful generation of $a$ demonstrates effective copy-paste behavior by extracting $a$ from the haystack and copying it over to the output.
We say an attention head $h$ copies a token $t$ if, during the generation of $t$ in the answer, $h$ assigns the maximum attention score to the same token $t$ in the needles. To quantify this behavior, the retrieval score of an attention head $h$ is defined as the fraction of tokens copied from $a$ by the head $h$ during decoding:
\begin{equation}
\vspace{-0.5em}
\mathrm{Retrieval\_Score}(h) = \frac{|g_h \cap a|}{|a|},
\end{equation}
where $g_h$ denotes the set of tokens copied by head $h$ to the output. Attention heads with the highest retrieval scores are selected as retrieval heads.

\paragraph{Shortcomings.} The scoring mechanism described above focuses only on attention heads that perform strict copy-paste operations, potentially missing heads involved in semantic-based retrieval, such as paraphrasing or reasoning over relevant context. Moreover, recent work has shown that heads identified through copy-paste metrics exhibit limited cross-domain generalizability \cite{zhao2024understandingsyntheticcontextextension}. This suggests that the simplified formulation may not fully capture the complexity of in-context retrieval behavior in LLMs and has limited relevance for downstream applications.

\section{{\headname}: Identifying Query-Focused Retrieval Heads}
\label{sec:method_qrhead}

    In this section, we introduce a new approach for detecting retrieval heads that significantly improves upon prior retrieval head detection. For clarity, we refer to our heads as  \underline{Q}uery-Focused \underline{R}etrieval Head (\headname{}) and the original retrieval head as \worsehead{}. See Figure~\ref{fig:methods} (bottom), our approach introduces two key improvements. First, we propose a query-focused retrieval score (QRscore), which captures query-context attention rather than relying solely on copy-paste behavior (§\ref{sec:qrscore}). Second, we leverage realistic tasks that require in-context retrieval to identify effective heads (§\ref{sec:qrdetection}).
We also present a comparison between \headname{} and \worsehead{} (§\ref{sec:comparison}).

\paragraph{Task formulation: LMs for in-context retrieval.} 
Our study focuses on the task of in-context retrieval with LMs, i.e., identifying relevant information from given context. Formally, let $\Sigma$ denote the vocabulary. Given an input query $q \in \Sigma^*$ and a context $D \in \Sigma^*$, the objective is to retrieve the most relevant information from the context with respect to $q$, denoted as $D_{[q]} \subseteq D$. Typically, the context $D$ consists of a sequence of passages (or chunks), represented as $D = \{d_1, d_2, \dots, d_N\}$. With both $q$ and $D$ jointly fed into an LM as input, we assign a score $\retscore(q, d_i)$ to each passage $d_i$ with respect to $q$. We measure the effectiveness of the retriever by evaluating whether the top-scored passages align with the ground-truth relevant documents $D^*_{[q]}$.\footnote{We note NIAH task can also be viewed as a special case of this formulation, where the gold document set only contains one document (the needle).}

\subsection{Scoring Heads with Query-Context Attention}
\label{sec:qrscore}

Instead of scoring attention heads based on their activations in copy-paste operations, we propose to evaluate them based on their effectiveness in realistic in-context retrieval tasks. This offers a more general and realistic measure of retrieval capability, as it captures semantic relevance rather than relying solely on verbatim copying.

\paragraph{Query-focused retrieval score (QRscore).} We use QRscore as a measure of the retrieval capability of an attention head in response to a specific query. Formally, 
let $h\in\mathcal{H}$ be an attention head within the language model, and let $A_h$ denote the attention weights (post-softmax) of head $h$ over a query prompt $\{D,q\}$, such as a prompt with a book followed by a question over its contents. The query-focused attention scores of head $h$ towards a document $d_i$ is calculated as follows:

\footnotesize
\begin{equation}
\label{eqn:score_perdoc}
\mathrm{QRscore}_h(q,d_i) =\frac{1}{|q|}\sum_{t_q\in q}\sum_{t_d \in d_i}A_h^{ t_q\rightarrow t_d},
\end{equation}
\normalsize

\noindent where $t_q$ denotes tokens in the query $q$, \(t_d\) represents tokens in the document \(d_i\), and \(A_h^{t_q\rightarrow t_d}\) is the attention weight of $h$ from \(t_q\) to \(t_d\).  This formulation quantifies the degree to which head \(h\) focuses on document \(d_i\) in response to $q$. Lastly, we aggregate the scores for all documents $d_i$ within the gold document set $D^*_{[q]}$, resulting in the final QRscore for head $h$ with respect to the query $q$:

\footnotesize
\begin{equation}
\label{eqn:score_agg}
\mathrm{QRscore}_{h}(q) =\frac{1}{|q|}\sum_{d_i\in D^*}\sum_{t_q\in q}\sum_{t_d \in d_i}A_h^{t_q\rightarrow t_d}
\end{equation}
\normalsize

\subsection{Detecting \headname{} on Real-World Tasks}
\label{sec:qrdetection}

With the QRscore defined in Eq.~\ref{eqn:score_agg}, we can now quantify the retrieval capabilities of each attention head over a given set of documents in response to a query. To achieve this, we leverage a head detection dataset $\mathcal{T} = \{(q, D, D^*_{[q]})\}$, which consists of a query $q$, a set of candidate documents $D$, and the corresponding gold documents $D^*_{[q]}$. Notably, our approach does not require explicit answers to the queries—only the annotations of the gold document. Using this detection dataset $\mathcal{T}$, we compute the effectiveness of an attention head $h$ for retrieval as follows:

\footnotesize
\begin{equation}
\mathrm{QRscore}_{h,\mathcal{T}}=\frac{1}{|\mathcal{T}|}\sum_{(q,D,D^*)\in\mathcal{T}}\mathrm{QRscore}_{h}(q)
\end{equation}
\normalsize

 \noindent As shown in Figure~\ref{fig:methods},  instead of synthetic needle-in-a-haystack task (NIAH)~\cite{niah}, we use more realistic in-context retrieval task for head detection (e.g., claim verification over books). We argue that more natural and realistic distractors provide more effective supervision that allows identifying heads that are better at differentiating relevant context from distracting context. We also note that even a small amount (< 100) of realistic data points can be sufficient, allowing us to find \headsname{} heads that contribute to improved downstream performance.

\subsection{Comparing {\headname{}} and Original Retrieval Head}
\label{sec:comparison}
We have introduced our method for detecting \headname{}. Here, we compare the \headname{} with original retrieval head (\worsehead{}) within the same model, using Llama-3.1-8B-Instruct~\cite{dubey2024llama} as a case study.

First, following the analysis setup of \citet{wu2025retrieval}, we measure the impact of pruning by the \textit{performance drop on NIAH test}. Specifically, we prune the top 32 heads (roughly 3\% of all attention heads in LLaMA-3.1-8B), following the commonly reported 5\% sparsity level of retrieval heads in \citet{wu2025retrieval,zhao2024understandingsyntheticcontextextension}.
As shown in Figure~\ref{fig:niah_teaser}, pruning the top 32 \headname{} results in near-complete failure on the NIAH performance, whereas pruning the top 32 \worsehead{} yields a much smaller performance decline.\footnote{See Appendix~\ref{appendix:qwen_niah} for results on Qwen-2.5-7B-Instruct.} In addition, we find \textbf{substantial divergence} between the two sets. Among the top 32 and top 64 heads, only 8 and 32 overlap, respectively. This less than 25\% overlap in the top 32 highlights the distinct roles of \headname{} and \worsehead{}.

\section{Using \headname{} to Build A General-Purpose Retriever}
\label{sec:qrretriever_method}

In this section, we describe how the detected \headname{} can be used in downstream applications. Specifically, we find the attention mass of \headname{} provides highly reliable signals for in-context retrieval. 

\subsection{The Method}
Given a selected set of \headname{} $\mathcal{H}^{\text{select}}$, a query $q$, and a collection of passages $D$, we compute the retrieval score for each passage $d_i$ by aggregating the QRscore across all heads in $\mathcal{H}^{\text{select}}$:
\begin{equation}
\small
\mathcal{R}(q,d_i)=\frac{1}{|\mathcal{H}^{\text{select}}|}\sum_{h\in\mathcal{H}^{\text{select}}}\mathrm{QRscore}_h(q,d_i).
\end{equation}
Passages are then ranked using their retrieval scores. We call our retrieval system \methodname{}. It offers several advantages: 
(1) \emph{General-purpose:} applicable across diverse domains without training,  unlike traditional retrievers that often are often limited in generalizing out of domain
(2) \emph{Model-agnostic:} compatible with any transformer-based LMs without modification, 
(3) \emph{Efficient:} leverages attention patterns to process long context simultaneously without expensive generation or pairwise comparisons.

\paragraph{Calibration.}
To mitigate intrinsic biases in LMs' attention weights, we adopt the score calibration method proposed by \citet{chen2025icr}. Instead of directly using $R(q,d_i)$ as the score, we additionally compute baseline score,  $R(q_{null},d_i)$, using a context-free null query $q_{null}$ (\texttt{"N/A"}). For each $d_i$, we use calibrated the score $R(q,d_i) - R(q_{null},d_i)$ as the final retriever score.

\begin{table*}[t]
    \centering
    \renewcommand{\tabcolsep}{1.2mm}
    \fontsize{7.75}{7.75}\selectfont
    \begin{tabular}{lcccccccccccccc}
    \toprule
    && \multicolumn{5}{c}{\textbf{LongMemEval}} && \multicolumn{5}{c}{\textbf{\textsc{Clipper}}} \\
     \cmidrule{3-7} \cmidrule{9-13}
    \multirow{3}{*}{\textbf{\textsc{Retriever}}} && \multicolumn{2}{c}{\textbf{\textsc{Retrieval}}}  && \multicolumn{2}{c}{\textbf{\textsc{End-to-End}}} && \multicolumn{2}{c}{\textbf{\textsc{Retrieval}}}  && \multicolumn{2}{c}{\textbf{\textsc{End-to-End}}} \\
     && \multicolumn{2}{c}{\textbf{\textsc{Recall@k}}}  && \multicolumn{2}{c}{\textbf{\textsc{Performance}}} && \multicolumn{2}{c}{\textbf{\textsc{Recall@k}}}  && \multicolumn{2}{c}{\textbf{\textsc{Performance}}} \\
        
         && k = 5 & k = 10 && Top-5 & Top-10 && k = 3 & k = 5 && Top-3 & Top-5 \\
        \cmidrule{1-1} \cmidrule{3-4} \cmidrule{6-7} \cmidrule{9-10} \cmidrule{12-13}
        \textbf{Base LM: Llama-3.2-3B-Instruct} \\
        Full context && - & - && \multicolumn{2}{c}{28.1} && - & - && \multicolumn{2}{c}{25.2} \\
         BM25 && 57.5 & 67.5 && 46.1 & 44.9 && 74.6 & 83.7 && 20.0 & 22.8 \\
         Contriever && 62.7 & 79.2 && \bf 48.6 & 46.5 && 60.2 & 78.9 && 12.6 & 18.4 \\
         Stella && 63.9 & 77.6 && 44.9 & \bf 47.7 && 83.3 & 90.0 && 21.3 & 25.1 \\
         \cmidrule{1-1}
          RankGPT && 1.8 & 3.4  && 23.5 & 23.3 && 16.8 & 27.3 && 3.6 & 8.8 \\
         RankGPT\textsuperscript{Bubble}  && 2.1 & 3.8 && 24.0 & 24.4 && 17.0 & 27.4 && 3.8 & 8.8 \\
         ICR  && 68.7 & 78.8 && 46.5 & 45.1 && 72.8 & 83.6 && 19.4 & 23.6 \\
         \methodname{} (Ours) && \bf 77.6 & \bf 86.6 && 47.4 & \bf 47.7 && \bf 85.5 & \bf 93.4 && 23.4 & \bf 26.9 \\
         \midrule
        \textbf{Base LM: Llama-3.1-8B-Instruct} \\
        Full context && - & - &&  \multicolumn{2}{c}{46.5} && - & - && \multicolumn{2}{c}{31.3} \\
         BM25 && 57.5 & 67.5 && 48.8 & 50.9 && 74.6 & 83.7 && 37.9 & 37.9 \\
         Contriever && 62.7 & 79.2 && 52.6 & 55.4 && 60.2 & 78.9 && 28.2 & 31.1 \\
         Stella && 63.9 & 77.6 && 50.9 & 58.4 && 83.3 & 90.0 && 38.8 & 39.6 \\
        \cmidrule{1-1}
        RankGPT && 2.1 & 4.0 && 26.7 & 24.2 && 30.0 & 39.4 && 15.9 & 19.4 \\
        RankGPT\textsuperscript{Bubble} && 8.3 & 9.0 && 28.1 & 27.0 && 36.7 & 44.3 && 19.7 & 20.4 \\
        ICR && 77.0 & 84.4 && 59.3 & 56.1 && 89.3 & 94.7 && 43.8 & \bf 42.5 \\
        \methodname{} (Ours) && \bf 85.5 & \bf 91.7 && \bf 59.8 & \bf 60.2 && \bf 93.8 & \bf 96.9 && \bf 47.6 & 41.9 \\
         \midrule
        \textbf{Base LM: Llama-3.1-70B-Instruct} \\
         Full context && - & - && \multicolumn{2}{c}{34.2} && - & - && \multicolumn{2}{c}{63.9} \\
         BM25 && 57.5 & 67.5 && 52.8 & 53.0 && 74.6 & 83.7 && 60.1 & 66.5 \\
         Contriever && 62.7 & 79.2 && 53.7 & 60.5 && 60.2 & 78.9 && 38.5 & 49.7 \\
         Stella && 63.9 & 77.6 && 56.3 & 62.3 && 83.3 & 90.0 && 65.9 & 71.2 \\
         \cmidrule{1-1}
         RankGPT && 1.8 & 3.5 && 21.2 & 27.4 && 57.0 & 63.4 && 44.7 & 50.4 \\
         RankGPT\textsuperscript{Bubble} && 47.9 & 49.0 && 44.0 & 42.6 && 74.3 & 78.8 && 58.4 & 61.5 \\
         ICR && 32.1 & 46.5 && 36.5 & 39.8  && 86.2 & 93.1 && 68.9 & 71.6 \\
         \methodname{} (Ours) && \bf 80.4 & \bf 88.5 && \bf 66.7 & \bf 67.7 && \bf 95.0 & \bf 98.2 && \bf 74.2 & \bf 73.3 \\
    \bottomrule
    \end{tabular}
    \caption{Results on LongMemEval and \clipper{}. The base model denotes the LM used for both the retriever and end-to-end generation. \headsname{} used for \clipper{} are found through using LongMemEval.}
    \label{tab:longmemeval}
\end{table*}

\subsection{Applications}

\paragraph{Long-context reasoning.} 
Long-context language models often struggle with performance degradation when processing long context~\cite{helmet,Longproc,liu-etal-2024-lost}. 
To address this, we integrate \methodname{} within a retrieval-augmented generation (RAG) framework.  Given a long-context input and a query, we segment the input into smaller chunks and use \methodname{} to score and subsequently extract the most relevant ones. The extracted context are concatenated to create a reduced context, that is then given to the LM for generating the final answer in another forward pass.

\paragraph{Passage re-ranking.}
Text retrieval powers many retrieval-augmented downstream applications \cite{lewis2020retrieval}.
A critical component in the retrieval pipeline is the re-ranker, which re-orders the passages returned by a first-stage retriever to enhance top passage relevance \citep{nogueira2020passagererankingbert,ma2024rankllama}.
\methodname{} can naturally be used as a re-ranker as part of any retrieval pipeline without any fine-tuning by simply concatenating the retrieved passages in the input and scoring their relevance directly.

\section{Experiments}
\label{sec:main_exp}

We evaluate \methodname{} on two tasks: long-context reasoning (\S\ref{sec:main_longctx}) and re-ranking (\S\ref{sec:main_beir}).

\subsection{Base Models and Baselines}
\paragraph{Base LMs.} We experiment with open-weight, instruction-tuned LMs from two families across different sizes, including Llama-3.2 (3B), Llama-3.1 (8B and 70B) of Llama family~\cite{dubey2024llama}, and Qwen2.5 (7B) of Qwen family~\cite{Qwen2TR}. With \methodname{}, we use 16 heads for models with fewer than 10B parameters, and 32 heads for Llama-3.1-70B. This corresponds to approximately 1–2\% of the total attention heads, given the sparsity of retrieval heads.

\paragraph{Baselines.} 
We compare our methods against several strong baselines. Following \citet{wu2025longmemeval}, we compare against dense retrievers, including Contriever~\cite{izacard2022contriever} and 1.5B Stella V5~\cite{zhang2025stella}, two popular strong dense retrievers. For Contriever, we truncate the input to $512$ tokens according to its maximum context length. We also compare against existing LLM-based re-rankers, including:
\begin{itemize}[noitemsep,leftmargin=10px]
    \item \textbf{RankGPT}~\cite{sun2024rankgpt} is a generative re-ranker that instructs LLMs to output the ranking order of a given set of documents based on a query. We experiment with two variants of RankGPT: (1) \textbf{RankGPT without sliding window}, which directly inputs all documents into the model prompt simultaneously, and (2) \textbf{RankGPT with sliding window} (RankGPT\textsuperscript{Bubble}), which leverages bubble sort to rank smaller subsets of documents incrementally.
    \item \textbf{In-Context-Re-ranking} ~\cite[ICR;][]{chen2025icr} is a re-ranker that also leverages the attention for relevance scoring. ICR uses full attention heads for scoring relevace, whereas we only use the attention weights of selected \headsname{}.
\end{itemize}

\begin{table*}[t]
    \centering
        \renewcommand{\tabcolsep}{1.2mm}
    \fontsize{7.5}{7.5}\selectfont
    \begin{tabular}{lcccccccccccc}
    \toprule
& \bf NQ & \bf COVID & \bf NFCorpus & \bf FiQA & \bf Scifact & \bf Scidocs & \bf FEVER & \bf Climate & \bf DBPedia & \bf Robust04 & \bf News & \bf Avg \\
\midrule
BM25 & 30.5 & 59.5 & 32.2 & 23.6 & 67.9 & 14.9 & 65.1 & 16.5 & 31.8 & 40.7 & 39.5 & 38.4 \\
\midrule
& \multicolumn{11}{c}{\textit{Base LM: Llama-3.2-3B-Instruct }}  \\
\cmidrule{0-12}
RankGPT & 30.0 & 59.5 & 32.2 & 23.6 & 67.9 & 14.9 & 65.9 & 17.1 & 31.8 & 40.7 & 39.5 & 38.5 \\
RankGPT\textsuperscript{Bubble} & 33.2 & 61.8 & 32.0 & 22.4 & 66.1 & 14.8 & 65.8 & 17.1 & 34.8 & 40.5 & 40.2 & 39.0 \\
\cmidrule{1-1}
ICR & 49.2 & 72.3 & 33.8 & 31.8 & 73.3 & 17.4 & 82.6 & 24.2 & 34.7 & 47.2 & 44.7 & 46.5 \\
{\methodname{} (Ours)} & \bf 54.9 & \bf 77.4 & \bf 35.1 & \bf 35.1 & \bf 74.7 & \bf 18.3 & \bf 83.7 & \bf 24.5 & \bf 36 & \bf 49.7 & \bf 45.1 & \bf 48.6 \\
\midrule
& \multicolumn{11}{c}{\textit{Base LM: Llama-3.1-8B-Instruct }}  \\
\cmidrule{0-12}

RankGPT & 30.0 & 59.5 & 32.2 & 23.6 & 67.9 & 14.9 & 65.9 & 16.8 & 31.8 & 40.7 & 39.5 & 38.4 \\
RankGPT\textsuperscript{Bubble} & 53.7 & 75.5 & 34.3 & 31.4 & 69.3 & 17.4 & 67.5 & 23.8 &  
\bf42.9 & 47.8 & \bf  46.2 & 46.3 \\
\cmidrule{1-1}
ICR & 54.0 & 73.3 & 34.8 & 35.6 & 75.5 & 19.0 & \bf 85.8 & \bf 24.8 & 36.9 & 49.0 & 44.5 & 48.5 \\
{\methodname{} (Ours)} & \bf 58.6 & \bf 77.5 & \bf 35.3 & \bf 39.1 & \bf 76.2 & \bf 19.4 & 85.3 & 23.9 & 37.2 & \bf 51.4 & \bf 46.2 & \bf 50.0 \\
\midrule
 
& \multicolumn{11}{c}{\textit{Base LM: Qwen-2.5-7B-Instruct }} \\
\cmidrule{0-12}
RankGPT & 30.0 & 59.5 & 32.2 & 23.6 & 67.9 & 14.9 & 65.9 & 16.8 & 31.8 & 40.7 & 39.5 & 38.4 \\
RankGPT\textsuperscript{Bubble} & 42.7 & \bf 70.5 & \bf 34.1 & \bf 29.5 & 69.3 & \bf 16.6 & 70.5 & 19.7 & \bf 37.1 & \bf 46.4 & \bf 43.6 & 43.6 \\
\cmidrule{1-1}
ICR & 43.1 & 66.1 & 32.7 & 27.0 & \bf 71.1 & 16.4 & 79.2 & 19.6 & 35.3 & 43.0 & 40.0 & 43.0 \\
{\methodname{} (Ours)} & \bf 49.9 & 67.7 & 33.1 & 29.2 & 71 & 15.3 & \bf 80.7 & \bf 20.1 & 35.7 & 43.7 & 39.8 & \bf 44.2 \\
\midrule
& \multicolumn{11}{c}{\textit{Base LM: Llama-3.1-70B-Instruct }}  \\
\cmidrule{0-12}
RankGPT & 45.4 & 62.7 & 33.6 & 28.6 & 71.3 & 16.1 & 74.2 & 18.9 & 37.6 & 41.3 & 39.8 & 42.7 \\
RankGPT\textsuperscript{Bubble} & 58.4 & \bf 81.2 & \bf 36.1 & 41.0 & 76.1 & 20.2 & 80.0 & \bf 25.1 & \bf 45.5 & \bf 59.0 & \bf 48.5 & \bf 51.9 \\
\cmidrule{1-1}
ICR & 57.9 & 72.3 & 34.2 & 38.9 & 74.6 & 19.4 & \bf 86.4 & 22.0 & 38.3 & 42.6 & 40.4 & 47.9 \\
{\methodname{} (Ours)} & \bf 62.1 & 73.9 & 34.7 & \bf 43.8 & \bf 76.8 & \bf 20.4 & 85.9 & 23.1 & 35.7 & 51.6 & 43.5 & 50.1 \\
\midrule
\rowcolor{gray!20}
& \multicolumn{11}{c}{\textit{Dual Encoder and Cross Encoder }} & \\
\cmidrule{0-12}
\rowcolor{gray!20}
Contriever & 44.6 & 67.5 & 32.8 & 28.4 & 67.1 & 18.9 & 64.2 & 28.0 & 39.5 & 45.7 & 41.7 & 43.5 \\
\rowcolor{gray!20}
GTR-t5-base & 51.4 & 74.8 & 32.5 & 34.7 & 62.1 & 15.8 & 72.9 & 26.8 & 37.1 & 46.1 & 42.8 & 45.2 \\
\rowcolor{gray!20}
BGE-Reranker-base & 55.2 & 66.4 & 31.0 & 31.7 & 70.8 & 15.7 & 88.6 & 36.5 & 42.5 & 39.9 & 37.0 & 46.8 \\
\rowcolor{gray!20}
msmarco-MiniLM & 55.8 & 74.3 & 35.2 & 35.1 & 68.5 & 17.5 & 80.4 & 25.5 & 45.3 & 47.9 & 43.0 & 48.0 \\
\bottomrule
    \end{tabular}
    \caption{Performance comparison (nDCG@10) on BEIR benchmarks across LMs. \methodname{} generally outperforms other baselines across all models. With Llama-3.1-70B, \methodname{} underperforms RankGPT with (Bubble sort), which requires substantial amount of LLM generation calls. We also report the performance of popular dual encoders and cross encoders for reference (gray box).}\vspace{-1em}
    \label{tab:beir_main}
\end{table*}

\subsection{Long-Context Multi-Hop Reasoning}
\label{sec:main_longctx}

\paragraph{Datasets.} We use 1) \textbf{LongMemEval}~\cite{wu2025longmemeval}, which evaluates the long-term memory capabilities of LLM-driven chat assistants, and 2) \textbf{CLIPPER}~\cite{pham2025clipper}, which evaluate claim-verification over books.
Both datasets feature long-context (90K to 120K) and require multi-hop reasoning over several pieces of evidences. We segment each dataset according to its natural structure (e.g., message in multi-turn conversation or chapters in a book). For evaluation, we measure retrieval performance using recall and assess downstream task performance with accuracy. Please refer to Appendix~\ref{app:eval_detail} for more details.

\paragraph{Data for head detection.}
We detect \headname{} using a small subset of data from LongMemEval, specifically the single-session-user subset (which only requires single-hop reasoning) consisting of 70 examples, which we exclude from downstream evaluation. We use the set of heads for both LongMemEval and CLIPPER, testing generalization to multi-hop reasoning.

\paragraph{\methodname{} achieves strong retrieval performance for long context, leading to improved end-to-end performance.} Table~\ref{tab:longmemeval} demonstrates the strong performance of \methodname{} on both LongMemEval and CLIPPER: it outperforms other baselines regarding both retrieval recall and end-to-end performance. For instance, Llama-3.1-8B-Instruct as the base LM, we see end-to-end performance improvements of over 10\% on both tasks with Llama-3.1-8B-Instruct.

\paragraph{\methodname{} generalizes across domains.}  \methodname{} outperform off-the-shelf dense retrievers (Contriever and Stella) by a large margin on LongMemEval and CLIPPER. In particular, none of these methods are trained or calibrated on CLIPPER. The better performance of \methodname{} suggests its stronger cross-domain generalization capabilities than dense retrievers.

It is worth noting that all test questions in LongMemEval are multi-hop, yet \methodname{} performs well on them despite only using single-hop questions to detect \headsname{}.

\paragraph{\methodname{} scales with model sizes.} We note that LM-based re-rankers show inconsistent performance patterns across model scales: RankGPT achieves near-zero retrieval recall with small models, and retrieval performance of ICR sees significant degradation when scaling up model size from 8B to 70B. At the same time, the performance of \methodname{} generally improves as the model size scales up.

\paragraph{Compact LMs exhibit strong retrieval capabilities despite their limited generation abilities.} As shown in Table~\ref{tab:longmemeval}, on LongMemEval, Llama-3.2-3B-Instruct achieves a Recall@10 of 86.6, closely matching the 88.5 score of the much larger LlamA-3.1-70B. However, Llama-3.2-3B only achieves a final end-to-end performance of 47.7, largely lagging 70B's performance of 67.7. We hypothesize that the long-context limitations of compact models stem more from their generation capabilities than from their retrieval abilities. These findings open up promising future directions. Compact LMs could serve as efficient long-context retrievers, paired with larger models for the actual generation.

\subsection{Passage Re-Ranking}

\label{sec:main_beir}
 To test the general applicability of \methodname{}, we evaluate our method on BEIR benchmark~\cite{thakur2021beir} consisting of diverse domains. We compare against zero-shot LLM-based re-rankers, RankGPT and ICR. We also report the performance of popular dual encoders and cross encoders from sentence transformer for references (please refer to Appendix~\ref{app:dual_and_cross_enc} for details about these models).
 
\paragraph{Setting.} Our setting largely follows prior work~\citep{chen2025icr}. We re-rank 200 passages retrieved using BM25, resulting a overall context length ranging from 16K to 64K depending on the average document length of domains.  We report the performance on the set of tasks used in \citet{chen2025icr}, we sub-sampled 512 random questions for each domain for evaluation.

\paragraph{Data for head detection.}  For BEIR, we utilize the 256 (held-out) data points from NQ and use them for on all other domains zero-shot.

\paragraph{Results.} Table~\ref{tab:beir_main} summarizes the BEIR results, demonstrating the strong effectiveness of \methodname{} as a general-purpose retriever. For models under 10B parameters, \methodname{} consistently outperforms other baselines. With Llama-3.1-8B, it achieves an average score of 50.0, outperforming RankGPT by 3.7 points and ICR by 1.5 points. For the larger Llama-3.1-70B model, \methodname{} significantly surpasses ICR, though it generally lags RankGPT\textsuperscript{Bubble} (which require over 200 generation calls). Nevertheless, \methodname{} achieves the best performance on several domains, such as SciFact and FiQA.
In addition, we find that \methodname{} directly benefits from the general-purpose capabilities of strong LLMs, outperforming popular dual encoders and cross-encoders that fine-tune various base models to improve retrieval performance.

\begin{table}[t]
    \centering
        \fontsize{7.5}{7.5}\selectfont
    \begin{tabular}{cl cc}
            \toprule
         \multicolumn{2}{l}{}  &  {\sc BEIR\textsubscript{shuffled}} &  {\sc LongMemEval} \\
            \multicolumn{2}{l}{}  &  {\sc nDCG@10} &  {\sc Recall} \\
         \midrule
         \multirow{4}{*}{\STAB{\rotatebox[origin=c]{90}{Llama-8B}}}
         & {\sc Random Heads} & 37.5 & 59.8 \\
         & {\sc Full Heads} & 42.8 &	77.0 \\
         &{\sc Retrieval Head} & 43.4	& 81.5 \\
         & {\headsname{}} & \bf 47.5 & \bf 85.5 \\
        \midrule
         \multirow{4}{*}{\STAB{\rotatebox[origin=c]{90}{Qwen-7B}}}
         & {\sc Random Heads} & 19.9	& 57.2 \\
         & {\sc  Full Heads} & 22.6	 & 67.1 \\
         &{\sc Retrieval Head} & 27.4	& 70.7 \\
         & {\headsname{}} & \bf 31.9& \bf 83.2 \\
         \bottomrule
    \end{tabular}
    \vspace{-0.5em}
    \caption{Comparison across head selection strategies. Using \headsname{} substantially outperforms using all heads or using original retrieval heads.}
    \label{tab:impact_head_selection}

\end{table}

\section{Analysis}
In this section, we analyze the key advantages of \headname{} (e.g., length generalization) and examine the factors underlying its effectiveness. Additional analyses on the impact of varying the number of heads (Appendix~\ref{appendix:num_heads}) and inference latency (Appendix~\ref{appendix:inference_time}) are provided in the appendix.

\label{sec:analysis}

\subsection{Impact of Head Selection}
\label{anlysis:impact_of_heads}
We provide further ablations on head selection, the core idea behind \methodname{}. We experiment with different sets of heads, including (1) using our \headsname{},  (2) using all the attention heads, (3) using original retrieval head, and (4) using randomly selected heads. We use 16 heads for all settings. Table~\ref{tab:impact_head_selection} presents the retrieval performance on LongMemEval re-ranking performance on BEIR (aggregated across tasks).\footnote{Here, we use BEIR where input documents are randomly shuffled rather than ranked by BM25. This setup allows uniform evaluation of retrieval across the full context.} The performance gaps between different strategies demonstrate the importance of using the right heads for retrieval. Using original retrieval heads is effective, compared to using random heads or full heads. Using our improved \headsname{} consistently outperforms using original retrieval heads.

\begin{table}[t]
    \centering
    \fontsize{7.5}{7.5}\selectfont
    \begin{tabular}{lccccc}
    \toprule
    & \multicolumn{5}{c}{\textit{Model: LLama-3.1-8B-Instruct \vspace{0.1em}}}  \\
     & \multicolumn{2}{c}{NQ+FEVER} && \multicolumn{2}{c}{LongMemEval}\\
         & 32K & 128K && 32K & 128K  \\
         \cmidrule{2-3} \cmidrule{5-6}
        ICR &  66.7	& 56.5	&& 85.2 & 	78.2\\
        \methodname{}\textsuperscript{32K} & \bf 70.1	& 63.9	&& 89.2 &	85.2\\
        \methodname{}\textsuperscript{128K} & 68.8	& \bf 67.2	&& \bf 89.2	& \bf 85.6\\
    \midrule
     & \multicolumn{5}{c}{\textit{Model: Qwen-2.5-7B-Instruct \vspace{0.1em}}}  \\
          & \multicolumn{2}{c}{NQ+FEVER} && \multicolumn{2}{c}{LongMemEval}\\
         & 32K & 64K && 32K & 64K  \\
         \cmidrule{2-3} \cmidrule{5-6}
ICR &  40.0 & 17.4&&	83.4 &	67.1 \\
        \methodname{}\textsuperscript{32K} & 51.9&	25.3	&&\bf90.2&\bf	77.9 \\
        \methodname{}\textsuperscript{64K} & \bf54.1	&\bf29.1	&&90.1	&77.0\\
    \bottomrule
    \end{tabular}
        \vspace{-0.5em}
    \caption{Results on short-to-long generalization of \headsname{}{}. \headsname{} detected with relative short-context data can be used for retrieval on longer context.  }
    \label{tab:len_gen}
        \vspace{-1em}

\end{table}

\begin{table}[t]
    \centering
    \fontsize{7.5}{7.5}\selectfont
    \setlength{\tabcolsep}{4pt}
    \begin{tabular}{lccc}
    \toprule
   & \multirow{2}{*}{Data} & {\sc BEIR} &  {\sc LongMem} \\
    & & {\sc nDCG@10} &  {\sc Recall} \\
    \cmidrule{2-4}
 
     & \multicolumn{3}{c}{\textit{Model: LLama-3.1-8B-Inst \vspace{0.1em}}}  \\
    \headsname{} & NQ& \bf 47.5 &  83.9  \\
    \headsname{} & LongMem&  47.1 & \bf 85.6  \\
    \headsname{} &  NIAH  & 46.8 & 83.4 \\
    \sc Retrieval Head &  NIAH & 43.4 & 81.5 \\
    \cmidrule{2-4}
 
    & \multicolumn{3}{c}{\textit{Model: Qwen-2.5-7B-Inst \vspace{0.1em}}}  \\
    \headsname{}  & NQ & 31.9 & 80.2 \\
    \headsname{}  & LongMem & \bf 32.1 & \bf 83.2 \\
    \headsname{} & NIAH & 30.9 & 79.7  \\
    \sc Retrieval Head  & NIAH & 27.4 & 70.7  \\
    \bottomrule
    \end{tabular}
        \vspace{-0.5em}

    \caption{Analysis of factors contributing to improved head selection. Applying QRScore (\S\ref{sec:qrscore}) on NIAH results in more effective heads than the original retrieval heads. Using QRScore on realistic tasks yields the most effective head selection overall.}
    \label{tab:better_selection}
        \vspace{-1em}

\end{table}

\subsection{Generalizability Across Lengths}
We test the length generalization of \headsname{}: if we detect \headsname{} on relatively short context length (32K), can the heads generalize to longer context lengths (128K)?

We test such short-to-long generalization by controlling the number of documents in context. This results in datasets of different lengths ranging from 32K to 128K tokens.  We detect \headsname{} from both short and long datasets and test their performance on re-ranking tasks (using two representative subsets: NQ and FEVER) and LongMemEval. For Qwen-2.5-7B, we set the longer context length to 64K due to its original 32K limit. As shown in Table~\ref{tab:len_gen}, \headname{} detected using short-context data can generalize to longer-context settings, though heads detected from longer data generally yield better long-context performance.

\subsection{Ablations on Scoring Function and Task Selection for Head Detection}
In \S\ref{sec:method_qrhead}, we describe two key factors for head detection: using query-context attention objective, and using realistic data.
To assess the importance of these factors, we experiment with detecting heads on NIAH using QRscore (\S\ref{sec:method_qrhead}). As shown in Table~\ref{tab:better_selection}, applying QRscore on NIAH leads to improved performance compared to using the original retrieval heads detected from the same task. However, using realistic tasks such as NQ and LongMemEval with QRscore yields the best overall performance. These results highlight the importance of both the scoring method and head detection data.

\subsection{Sensitivity of \headsname{} Detection to Variation in Detection Data}

\begin{table}[t]
    \centering
    \fontsize{7.5}{7.5}\selectfont
    \begin{tabular}{lccccc}
    \toprule
 & \multicolumn{3}{c}{Overlap (Top 64) } & & BEIR  \\
 & Set0 & Set1 & Set2 & & nDCG@10 \\
 \cmidrule{2-4} \cmidrule{6-6}
 & \multicolumn{5}{c}{\textit{Model: LLama-3.1-8B-Inst \vspace{0.1em}}}  \\
\headsname{}\textsuperscript{Set0} & 64 & 51 & 51 && 49.8 \\
\headsname{}\textsuperscript{Set1} & 51 & 64 & 53 && 49.7 \\
\headsname{}\textsuperscript{Set2} & 51 & 53 & 64 && 49.9 \\
 \cmidrule{2-4} \cmidrule{6-6}
 & \multicolumn{5}{c}{\textit{Model: Qwen-2.5-7B-Inst \vspace{0.1em}}}  \\
\headsname{}\textsuperscript{Set0} & 64 & 50 & 53 && 44.2 \\
\headsname{}\textsuperscript{Set1} & 50 & 64 & 57 && 44.4 \\
\headsname{}\textsuperscript{Set2} & 53 & 57 & 64 && 44.5 \\
\bottomrule
    \end{tabular}
    \vspace{-0.5em}
    \caption{\textbf{Left}: Overlap in \headsname{} identified using three disjoint sets of 128 random samples from NQ.
\textbf{Right}: BEIR performance (nDCG@10) using \headsname{} detected from each sample set.}
    \label{tab:sensetivity_to_data}
\end{table}

In Section~\ref{sec:main_beir}, we show using a small number of samples from NQ is sufficient to identify effective \headsname{} for BEIR re-ranking tasks. We assess the robustness of this head detection process to different random samples of detection set, by experimenting with three disjoint random subsets of NQ, each containing 128 examples.
Table~\ref{tab:sensetivity_to_data} presents the overlap among the top-64 heads selected from these subsets and their performance on BEIR benchmark. Across two LLMs from different model families (Llama and Qwen), we observe a high degree of consistency with over 50 heads overlapping among the top 64 across subsets. Furthermore, the downstream performance remains stable across these variations. These results indicate that \methodname{} can be reliably identified using a small sample of data.

\section{Related Work}

\paragraph{LM-based retrieval and re-ranking.}
LMs are widely used in retrieval, including embedding-based methods~\cite{muennighoff2022sgpt,lee2021phraseretrievallearnspassage} and generative approaches~\cite{tay2022transformermemorydifferentiablesearch, decao2021autoregressiveentityretrieval, sun2023learningtokenizegenerativeretrieval}. For re-ranking, instruction-tuned LMs been adapted as re-rankers in various ways~\cite{sun2024rankgpt,drozdov2023paradepassagerankingusing,sachan2023improvingpassageretrievalzeroshot,ma2023zeroshotlistwisedocumentreranking,pradeep2023rankzephyreffectiverobustzeroshot}, leveraging their generation capabilities.
Similar to our approach, recent work has explored using logits~\cite{reddy2024firstfasterimprovedlistwise} or aggregated attention scores~\cite{chen2025icr} for re-ranking. In contrast, we identify a specialized set of attention heads responsible for retrieval, offering improved performance and interpretability.

\paragraph{Localizing model behavior.} 
Interpretability studies have shown that many core behaviors of LMs, including in-context learning~\cite{olsson2022incontextlearninginductionheads,todd2024function,mcdougall2023copysuppressioncomprehensivelyunderstanding} and retrieval~\cite{wu2025retrieval}, can be traced to specialized transformer modules~\cite{meng2022locating,dai-etal-2022-knowledge,stolfo2024confidence}. Techniques have been proposed to localize such modules with a small amount of data~\cite{meng2022locating,pmlr-v236-geiger24a,bhaskar2024finding}, and to intervene on them for control~\cite{li2023inferencetime,yin2024lofit,huang2025improvingcontextualfaithfulnesslarge} or efficiency~\cite{tang2025razorattention,xiao2025duoattention,liu2024slidingwindowsendexploring}. However, only a few works~\cite{zhao2024understandingsyntheticcontextextension} have examined attention head specialization in long-context settings, where attention is known to be not robust~\cite{liu-etal-2024-lost,xiao2024efficient}, and it is an open question if intervening the localized modules is crucial in practical settings \cite{hase2023does,wang2024does}.
Our work contributes to this line of research by finding better specialized set of attention heads that explain the model behavior for query-focused long-context retrieval, and that can be practically useful for zero-shot efficient retrieval.

\section{Conclusion}

We introduced \underline{Q}uery-Focused \underline{R}etrieval Heads (\headsname{}), a set of attention heads specialized in identifying query-relevant information in long-context inputs. Detected using query-context attention scores on realistic data, \headsname{} are better aligned with practical retrieval tasks than original retrieval heads. Built on top of \headsname{}, our retrieval method \methodname{} achieves strong performance on both long-context reasoning and re-ranking tasks, outperforming dense retrievers and other LLM-based re-rankers in many settings. These findings highlight the practical utility of \headsname{} and offer insights for further improving retrieval with LMs.

\section*{Limitations}
Our work detects improved retrieval heads and builds general-purpose retrievers based on them. We do not explore techniques that involve updating model parameters, as our goal is to develop flexible methods that can directly use off-the-shelf models as retrievers. Consequently, we leave to future work the investigation of parameter-updating techniques that leverage insights from \headsname{}.

While our method finds that \headsname{} can enhance downstream performance, and shows the importance of two factors leading to selection of better heads. We lack a complete understanding of the internal mechanism accounting for \headsname{}'s effectiveness. Future work could apply circuit analysis techniques (e.g., \citet{bhaskar2024finding, shi2024hypothesis}) to dissect the fine-grained behaviors and roles of these heads.

Our evaluation primarily targets passage re-ranking and long-context multi-hop reasoning tasks. Although our approach is conceptually applicable to broader long-context tasks—such as long-document summarization \cite{shaham2023, summaryhaystack}---it remains unclear whether it generalizes to such tasks without thorough empirical validation.

Finally, our experiments are limited to English datasets. As LMs may exhibit different behaviors across languages, the cross-lingual robustness of our approach remains an open question.

\section{Acknowledgments}
This work is gratefully supported by an NSF CAREER award (IIS-2239290) and a grant from Intel. Howard Yen is supported by the William A. Dippel’ 50 * 55 Graduate Fellowship.

\bibliography{custom}

\newpage
\appendix

\section{Details about Evaluation Datasets}
\label{app:eval_detail}
We use LongMemEval~\cite{wu2025longmemeval} and CLIPPER~\cite{pham2025clipper} for evaluating our systems on long-context reasoning.

\paragraph{LongMemEval} evaluates the long-term memory capabilities of LLM-driven chat assistants across five fundamental abilities: information extraction, multi-session reasoning, temporal reasoning, knowledge updates, and abstention. We segment the LongMemEval-S dataset ($\sim$115k tokens/question) at the round level, where each round is a document consisting of a single user message paired with the corresponding assistant response. 

\paragraph{CLIPPER} targets narrative claim verification—a challenging long-context reasoning task that requires verifying claims over entire books, with an average length of $90$K tokens and $23$ chapters. In CLIPPER, data is split at the chapter level, with each chapter treated as an individual document during retrieval. 

\paragraph{Evaluation Process} For each question, we first feed the entire context (e.g., all chapters or dialogue rounds) into the language model without using any first-stage retriever. We compute a retrieval score for each document or segment using our method described in §\ref{sec:qrretriever_method}. We then select the top-$k$ documents based the scores, concatenate them, and feed them together with the query into the language model in a second pass to generate the final answer. We choose $k=5,10$ for LongMemEval and $k=3,5$ for Clipper. We report retrieval performance using recall and downstream task performance using accuracy.

\begin{figure}[t]
    \centering
    \includegraphics[width=1.0\linewidth]{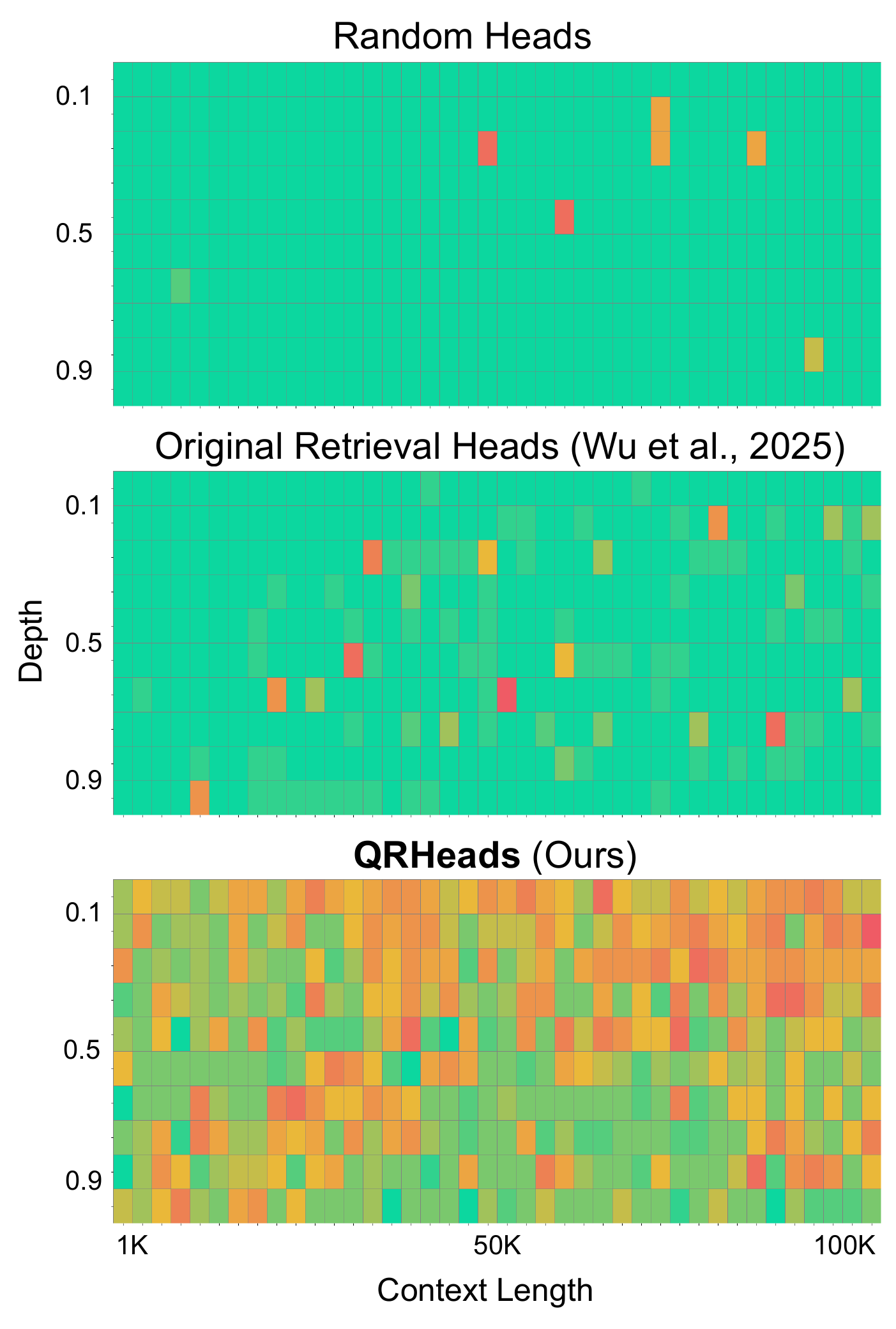}
    \caption{\textbf{Top}: Masking 16 random heads of Qwen2.5-7B-Instruct. \textbf{Middle}: Masking the top 16 original retrieval heads~\cite{wu2025retrieval}. \textbf{Bottom}: Masking the top 16 QRHeads.}
    \label{fig:niah_qwen_16}
\end{figure}

\begin{figure}[t]
    \centering
    \includegraphics[width=1.0\linewidth]{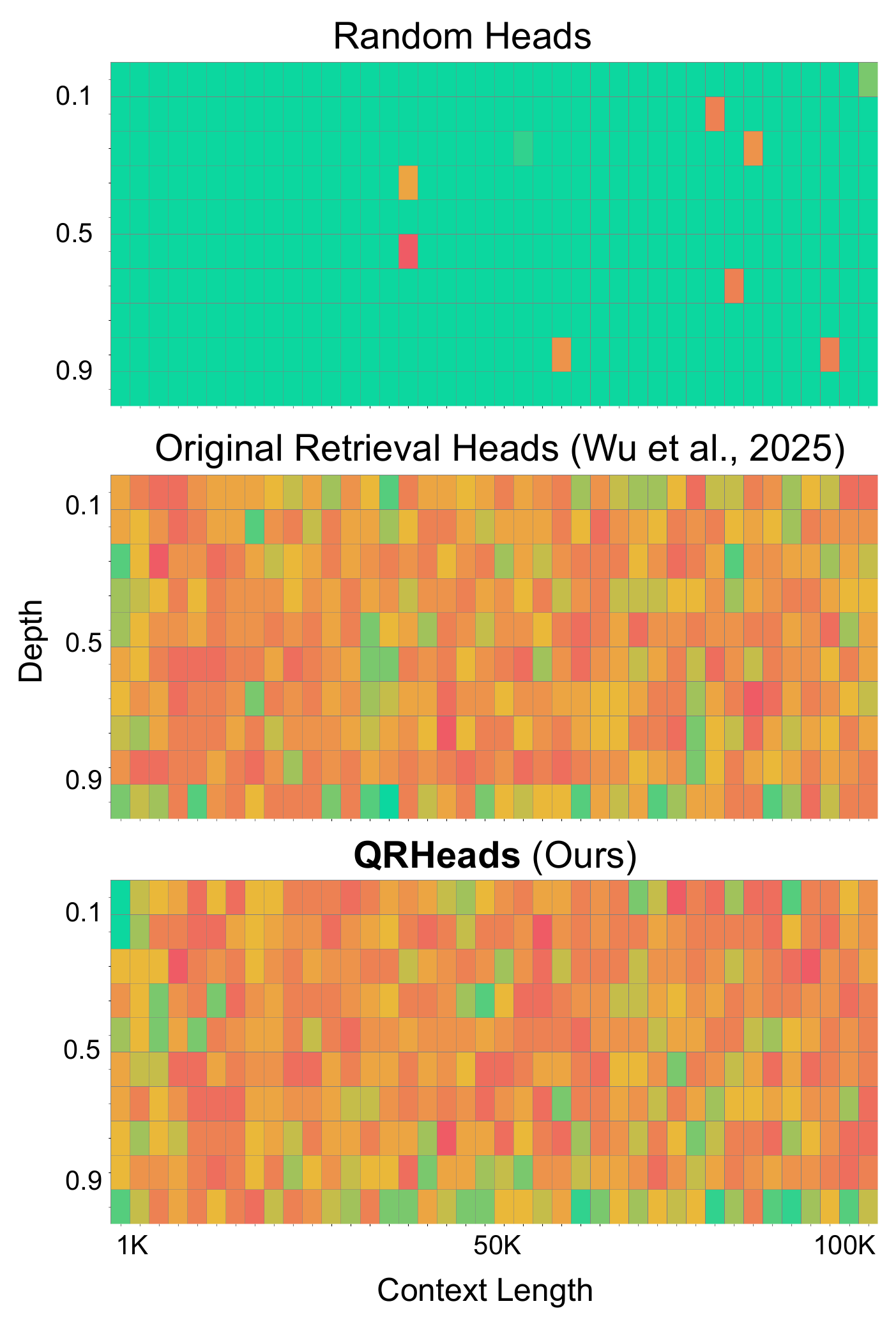}
    \caption{\textbf{Top}: Masking 32 random heads of Qwen2.5-7B-Instruct. \textbf{Middle}: Masking the top 32 original retrieval heads~\cite{wu2025retrieval}. \textbf{Bottom}: Masking the top 32 QRHeads.}
    \label{fig:niah_qwen_32}
\end{figure}

\section{NIAH Test on Qwen-2.5-7B-Instruct}
\label{appendix:qwen_niah}
We evaluate Qwen-2.5-7B-Instruct on the NIAH test by masking selected attention heads. As shown in Figure~\ref{fig:niah_qwen_16} and Figure~\ref{fig:niah_qwen_32}, pruning the top 16 \headname{} leads to a more substantial degradation in NIAH performance compared to pruning the top 16 \worsehead{}, indicating the greater functional importance of \headname{}. When pruning the top 32 heads, the performance gap between \headname{} and \worsehead{} narrows, suggesting that \headname{} achieves better efficiency and effectiveness with fewer heads for retrieval in NIAH task.

\section{Details about Dual Encoder and Cross Encoder Baselines on BEIR}
\label{app:dual_and_cross_enc}

In \S\ref{sec:main_beir}, we report the performance of several traditional retrievers, including dual encoders~\cite{karpukhin-etal-2020-dense} and cross-encoders~\cite{Yang2020IsRM}. We use popular models from the SentenceTransformers library~\cite{reimers-2019-sentence-bert}.\footnote{\url{https://sbert.net/}} Specifically, we include the following dual encoders:
\begin{itemize}
\item \textbf{Contriever}\cite{izacard2022contriever}: We use the checkpoint from \url{https://huggingface.co/facebook/contriever-msmarco}, which is fine-tuned on MS MARCO.
\item \textbf{GTR-T5-Base}\cite{ni-etal-2022-large}: We use the checkpoint from \url{https://huggingface.co/sentence-transformers/gtr-t5-base}.
\end{itemize}

We also include two cross-encoders fine-tuned on MS MARCO:
\begin{itemize}
\item \textbf{BGE-Reranker-Base}\cite{xiao2024c}: We use the checkpoint from \url{https://huggingface.co/BAAI/bge-reranker-base}.
\item \textbf{MSMARCO-MiniLM}\cite{reimers-2019-sentence-bert}: We use the checkpoint from \url{https://huggingface.co/cross-encoder/ms-marco-MiniLM-L6-v2}.
\end{itemize}

\section{Prompt Templates}
We provide prompt templates used in our experiments for ICR and \methodname{} in Figure \ref{fig:qr-prompt} and rankGPT in Figure \ref{fig:rankgpt-prompt}.

\begin{figure}[h!]
\centering
\begin{tcolorbox}[colframe=black, boxrule=0.8pt]
\small
\ttfamily
\{\texttt{prompt\_prefix}\} Here are some paragraphs:\\

[1] \{Title 1 (if available)\}\\  
\{Paragraph text 1\}\\

[2] \{Title 2 (if available)\}\\
\{Paragraph text 2\}\\

...\\

Please find information that is relevant to the following query in the paragraphs above.\\

Query: \{query\}\{prompt\_suffix\}
\end{tcolorbox}
\caption{Prompt used for ICR and \methodname{}.}
\label{fig:qr-prompt}
\end{figure}

\begin{figure}[h!]
\centering
\begin{tcolorbox}[colframe=black, boxrule=0.8pt]
\small
\ttfamily
\{\texttt{prompt\_prefix}\} This is an intelligent assistant that can rank passages based on their relevancy to the query.\\

The following are \{N\} passages, each indicated by a numbered identifier [i]. I can rank them based on their relevance to the query: "\{query\}"\\

[1] \{Title 1 (if available)\}\\  
\{Paragraph text 1\}\\

[2] \{Title 2 (if available)\}\\
\{Paragraph text 2\}\\

...\\

The search query is: "\{query\}". I will rank the \{N\} passages above based on their relevance to the search query. The passages will be listed in descending order using identifiers, the most relevant passages should be listed first and the output format should be [] > [] > etc, e.g., [1] > [2] > etc. Be sure to list all \{N\} ranked passages and do not explain your ranking until after the list is done. \{prompt\_suffix\} Ranked Passages: [
\end{tcolorbox}
\caption{Prompt used for rankGPT.}
\label{fig:rankgpt-prompt}
\end{figure}

\section{Inference Time Comparison between \methodname{}, ICR, and RankGPT}
\label{appendix:inference_time}
We compare the latency of \methodname{} with other LLM-based re-rankers (ICR and RankGPT). In Table \ref{tab:latency}, we report both latency and performance of Llama-3.1-8B and Llama-3.1-70B on NQ dataset. Compared to RankGPT, \methodname{} is significantly more time-efficient. This is primarily because (1) \methodname{} avoids autoregressive generation, and (2) it does not rely on bubble sort, which requires multiple rounds of generation to compare elements sequentially within each bubble.

\begin{table}[t]
    \centering
    \fontsize{8.5}{8.5}\selectfont
    \setlength{\tabcolsep}{4pt}
    \begin{tabular}{lcc}
    \toprule
   & \multirow{2}{*}{Avg Latency (s/sample)} & {\sc NQ} \\
    &  & {\sc nDCG@10} \\
    \cmidrule{2-3}
 
     & \multicolumn{2}{c}{\textit{Model: LLama-3.1-8B-Inst \vspace{0.1em}}}  \\
    RankGPT\textsuperscript{Bubble} & 14.5 &  53.7  \\
    ICR & 2.2 & 54.0  \\
    \methodname{} &  2.2 & 58.6 \\
    \cmidrule{2-3}
 
    & \multicolumn{2}{c}{\textit{Model: Llama-3.1-70B-Inst \vspace{0.1em}}}  \\
    RankGPT\textsuperscript{Bubble} & 63.9 & 58.4 \\
    ICR & 13.7 & 57.9 \\
    \methodname{} & 13.7 & 62.1  \\
    \bottomrule
    \end{tabular}

    \caption{Inference time and retrieval performance comparison on NQ.}
    \label{tab:latency}

\end{table}

\section{Determine the Number of Heads Used}
\label{appendix:num_heads}
We select the number of heads based on the sparsity level reported in prior work on retrieval heads \cite{wu2025retrieval}. In Table \ref{tab:head_subsets}, we include additional results on BEIR NQ using varying numbers of heads. The results show that performance remains consistently strong when using a small, top-ranked subset of heads, but degrades as more heads are included. This trend aligns with the sparsity property observed in retrieval heads. 

\begin{table}[t]
\centering
\fontsize{8.5}{8.5}\selectfont
\setlength{\tabcolsep}{6pt}
\begin{tabular}{lc}
\toprule
\textbf{Model / \#Heads} & \sc NQ (nDCG@10) \\
\midrule
\multicolumn{2}{l}{\textbf{Llama-3.1-8B} (\#heads = 1024)} \\
16 ($\sim$1.5\%)  & 58.6 \\
64 ($\sim$6\%)    & 57.4 \\
256 ($\sim$20\%)  & 56.6   \\
\midrule
\multicolumn{2}{l}{\textbf{Llama-3.1-70B} (\#heads = 5120)} \\
32 ($\sim$0.5\%)  & 62.1 \\
128 ($\sim$2\%)   & 61.9   \\
512 ($\sim$10\%)  & 61.0   \\
\bottomrule
\end{tabular}

\caption{Performance with different numbers of selected heads.}
\label{tab:head_subsets}
\end{table}

\section{License of Datasets}
The licenses datasets used in our work include:
\begin{itemize}
    \item LongMemEval~\cite{wu2025longmemeval} under MIT License.
    \item Clipper~\cite{pham2025clipper} under Apache license 2.0.
    \item NQ~\cite{kwiatkowski2019natural} under Creative Commons Attribution Share Alike 3.0.
    \item BEIR~\cite{thakur2021beir} under Creative Commons Attribution Share Alike 4.0 Read on choosealicense.com
\end{itemize}

\section{Computational Resources and Model Sizes}
We use Llama-3.2 (3B), Llama-3.1 (8B and 70B)~\cite{dubey2024llama}, and Qwen2.5 (7B)~\cite{Qwen2TR}. 8B models were run using a single NVIDIA A100 GPU with 80GB of memory, and 70B models were run using 4 A100 GPUs. All experiments were conducted on A100-based infrastructure.

\section{Potential Risks of Our Work}
N/A. Our work investigates the capabilities of existing language models, without proposing new model architectures or training procedures. While large language models pose well-known risks—including potential misuse, generation of harmful content, and encoding of societal biases—our study does not introduce new risks beyond those already covered in the broader literature. As such, we do not believe any specific risk mitigation measures are necessary for the scope of this work.

\end{document}